\documentclass{article}
\usepackage{arxiv}

\usepackage[utf8]{inputenc} %
\usepackage[T1]{fontenc}    %
\usepackage{hyperref}       %
\usepackage{url}            %
\usepackage{booktabs}       %
\usepackage{amsfonts}       %
\usepackage{nicefrac}       %
\usepackage{microtype}      %
\usepackage{lipsum}

\usepackage{booktabs}
\usepackage{bm}
\usepackage{xcolor}
\usepackage{tikz}
\usepackage{amsbsy}
\usepackage{amssymb, amsmath}
\usepackage{arydshln}
\usepackage{bbding}

\usepackage{xspace}

\makeatletter
\DeclareRobustCommand\onedot{\futurelet\@let@token\@onedot}
\def\@onedot{\ifx\@let@token.\else.\null\fi\xspace}

\def\eg{\emph{e.g}\onedot}

\makeatother

\title{Multi-Stage Fusion for One-Click Segmentation}

\author{
 Soumajit Majumder\\
 Institute of Computer Science II\\
 University of Bonn, Germany\\
 \And
 Ansh Khurana\\
 Dept. of Computer Science and Engineering \\
 Indian Institute of Technology Bombay
 \And
 Abhinav Rai\\ 
 School of Computing\\ National University of Singapore\\
 \And 
 Angela Yao\\
 School of Computing\\ National University of Singapore\\
  }

\begin{document}

\maketitle

\begin{abstract}
Segmenting objects of interest in an image is an essential building block of applications such as photo-editing and image analysis. Under interactive settings, one should achieve good segmentations while minimizing user input. Current deep learning-based interactive segmentation approaches use early fusion and incorporate user cues at the image input layer. Since segmentation CNNs have many layers, early fusion may weaken the influence of user interactions on the final prediction results. As such, we propose a new multi-stage guidance framework for interactive segmentation. By incorporating user cues at different stages of the network, we allow user interactions to impact the final segmentation output in a more direct way.  Our proposed framework has a negligible increase in parameter count compared to early-fusion frameworks. We perform extensive experimentation on the standard interactive instance segmentation and one-click segmentation benchmarks and report state-of-the-art performance.
\end{abstract}

\section{Introduction}
\label{sec:introduction}

\begin{figure}[t!]
	\begin{center}
	\includegraphics[width=1\linewidth]{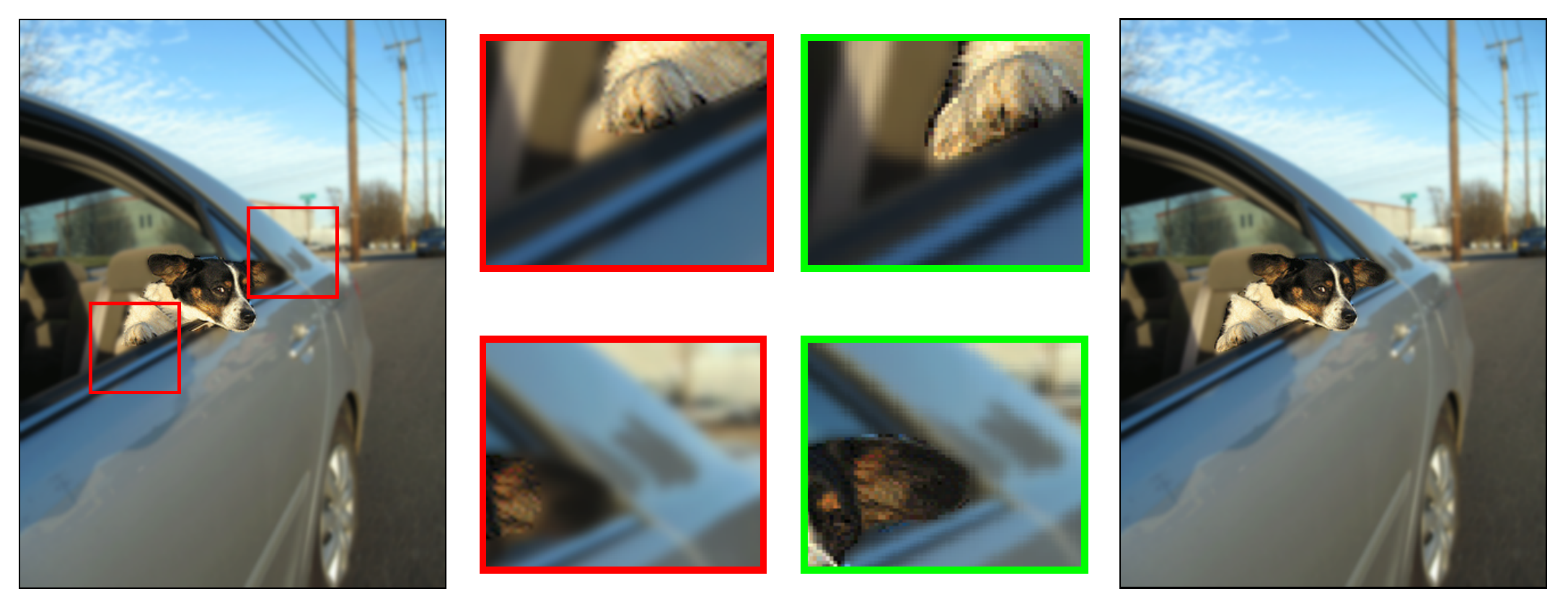}
	\end{center}
	\caption{\textbf{Motivation.} We consider the popular special-effect filter used in mobile photography - background blur. Here the user intends to blur the rest of the image barring the dog. In most existing interactive segmentation approaches\protect~\cite{ifcn,itis,majumder19}, the user click (here placed on the dog) is leveraged only at the input layer and its influence diminishes through the layers. This can result in unsatisfactory image effects, e.g portions of the dog's elbow and ear are wrongly classified as background and are mistakenly blurred (shown in enlarged red boxes). Our proposed multi-stage fusion allows user click to have a more direct effect leading to improvement in segmentation quality (shown in enlarged green boxes).}
	\label{fig:application}
\end{figure}

The widespread availability of smartphones had made taking photos easier than ever. In a typical image capturing scenario, the user taps the device touchscreen to focus on the object of interest. This tap directly locates the object in the scene and can be leveraged for segmentation. Generated segmentations are implicit, but are applicable for downstream photo applications, such as simulated `bokeh' or other special-effects filters such as background blur (see Fig.~\ref{fig:application}). In this work, we tackle~\emph{``tap-and-shoot segmentation''}~\cite{tapnshoot}, a special case of interactive instance segmentation. 

Interactive segmentation leverages inputs such as clicks, scribbles, or bounding boxes to help segment objects from the background down to the pixel level.  Two key differences distinguish tap-and-shoot segmentation from standard interactive segmentation. \emph{First}, tap-and-shoot uses only \emph{``positive''} clicks marking foreground, as we assume that the user clicks (only) on the object of interest during the capture process. Standard interactive segmentation uses both positive and negative clicks~\cite{ifcn,itis,majumder19} to respectively indicate the object of interest versus background or non-relevant foreground objects.  \emph{Secondly}, tap-and-shoot has a strong focus on maximizing the mean intersection over union (mIoU) with a single click because the target application is casual photography.  In contrast, standard interactive segmentation tries to achieve some threshold mIoU (\eg 85\%) while minimizing the total number of clicks.

This second distinction is subtle but critical for designing and learning tap-and-shoot segmentation frameworks.  Our finding is that existing approaches fare poorly with only one or two clicks -- they are simply not trained to maximize performance under such settings. To make the most of the first (few) click(s), we hypothesize that user cues' guidance should be fused into the network at multiple locations rather than via early fusion.  Just as gradients vanish towards the initial layers during back-propagation, input signals also diminish as it makes a forward pass through the network. The many layers of deep CNNs further exacerbate this effect~\cite{twostream,park2019}. A late fusion would allow the user interaction to have a direct and more pronounced effect on the final segmentation mask. To this end, we propose an interactive segmentation framework with multi-stage fusion and demonstrate its advantages over the common early fusion frameworks and other alternatives.  Specifically, we propose a light-weight fusion block that encodes the user click transformation and allows a shorter connection from user inputs to the final segmentation layer.

Most similar in spirit to our framework is~\cite{twostream} and~\cite{guidedprop}.  These two works also propose alternatives to early fusion but are extremely parameter heavy.  For example,~\cite{twostream} uses two dedicated VGG~\cite{vgg} networks to to extract features from the image and the user interactions separately before fusing into a final instance segmentation mask (see Fig.~\ref{fig:motivation}(c)). \cite{guidedprop} uses a single stream but applies a simple late fusion of element-wise multiplication on the feature maps (see Fig.~\ref{fig:motivation}(b)). It therefore has separate `positive' and `negative' feature maps and the number of weights for the following layer increases by a factor of $2$.  For VGG, this doubles the parameters of the ensuing \emph{`fc6'} layer from $100$ to $200$ million.  Compared to~\cite{guidedprop}, our last-stage fusion approach is light-weight and uses less than $1.5\%$ more trainable parameters.

Our contributions are summarized as follows: 
\begin{itemize}
    \item We propose a novel one-click interactive segmentation framework that fuses user guidance at different network stages. 
    \item We demonstrate that multi-stage fusion is highly beneficial for propagating guidance and increasing the mIoU since it allows user interaction to have a more direct impact on the final segmentation. 
    \item Comprehensive experiments on six benchmarks show that our approach significantly outperforms existing state-of-the-art for both tap-and-shoot and standard interactive instance segmentation.
\end{itemize}

\section{Related Works}

As an essential building block of image/video editing applications, interactive segmentation and dates back decades~\cite{scissors}. The latest methods~\cite{itis,ifcn,majumder19,twostream,guidedprop} integrate deep architectures 
such as FCN-8s~\cite{fcn} or DeepLab~\cite{deeplabv3,deeplabv2}. Most of these approaches integrate user cues in the input stage. The clicks are transformed into `guidance' maps and appended to the three-channel colour image input before being passed through a CNN~\cite{itis,ifcn,majumder19}. 

Early \textbf{Interactive Instance Segmentation} approaches used graph-cuts~\cite{graphcuts,grabcut}, geodesics, or a combination~\cite{geodesic}. These methods' performance is limited as they separate the foreground and background based on low-level colour and texture features. Consequently, for scenes where foreground and background are similar in appearance, or lighting and contrast is low, more labelling effort from the users to achieve good segmentations~\cite{ifcn}. Recently, deep convolutional neural networks~\cite{fcn,deeplabv3} have been incorporated into interactive segmentation frameworks. Initially,~\cite{ifcn} used Euclidean distance-based guidance maps to represent user-provided clicks and are passed along with the input RGB image through a fully convolutional network.

\begin{figure}[t!]
	\begin{center}
	\includegraphics[width=0.92\linewidth]{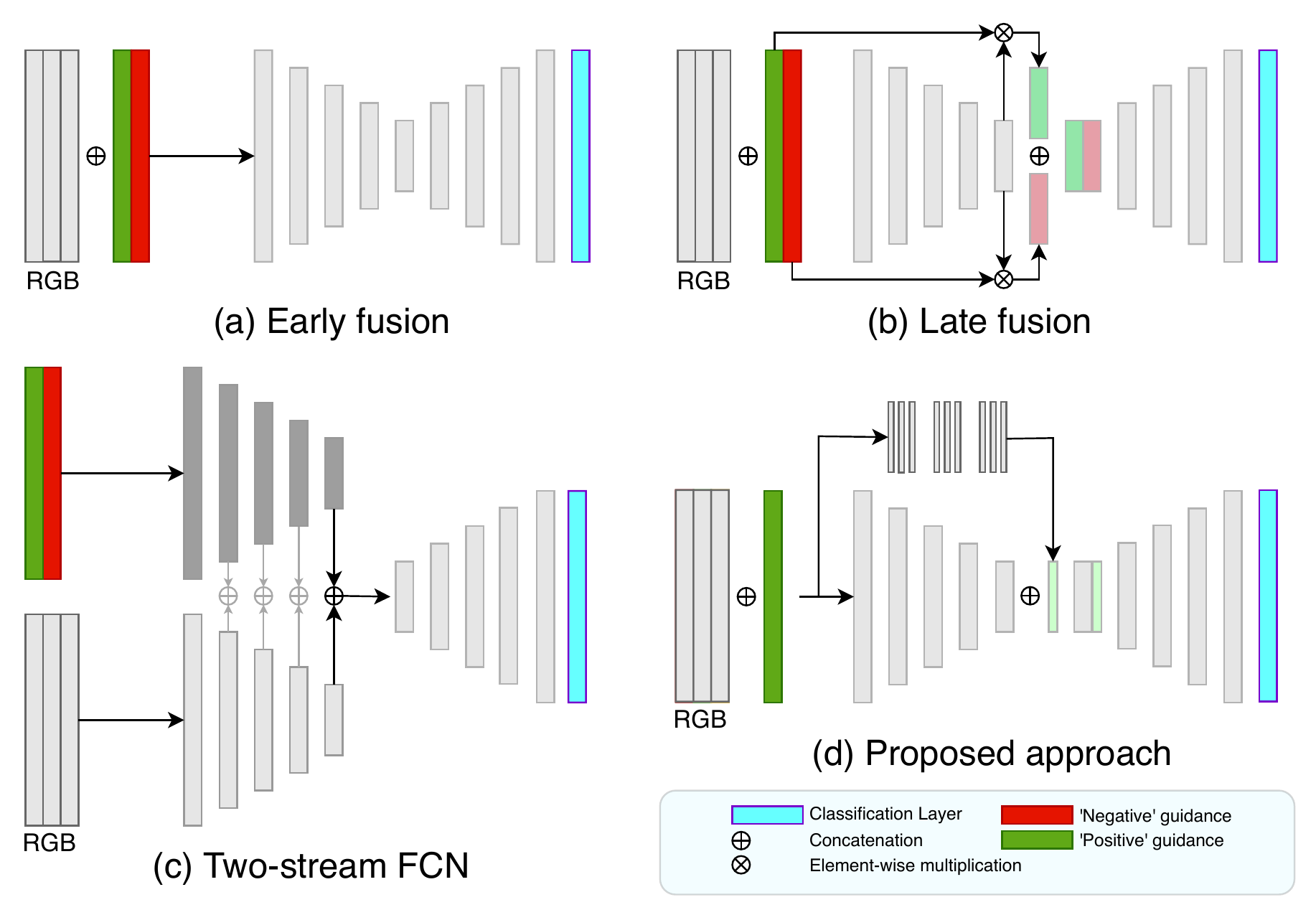}
	\end{center}
	\caption{
	(a) Existing interactive instance segmentation and ``tap-and-shoot" segmentation techniques concatenate user provided cues as an extra guidance map(s) (for `positive' and `negative' clicks) with the RGB and pass everything through a segmentation network. (b-c) Other alternative approaches are extremely parameter heavy. (b) The work of~\cite{twostream} uses two dedicated VGG~\cite{vgg} networks for extracting features from image and user interactions \emph{separately}. (c) The work of~\cite{guidedprop} performs late fusion via element-wise multiplication on the feature maps which requires an additional $100$ million parameters. (d) We leverage user guidance at the input (early fusion) and via late fusion. Our multi-stage fusion reduces the layers of abstraction and allows user interactions to have a more direct impact on the final output.
}
	\label{fig:motivation}
\end{figure}

Subsequent works made extensions with newer CNN architectures~\cite{itis}, iterative training procedures~\cite{itis} and structure-aware guidance maps~\cite{majumder19}. These works share a structural similarity: the guidance maps are concatenated with the RGB image as additional channels at the first (input) layer.  We refer to this form of structure as early fusion (see Fig.~\ref{fig:motivation}(a)). Architecture-wise, early fusion is simple and easy to train; however, user inputs' influence gets diminished through the layers. \\

\noindent\textbf{Tap-and-Shoot Segmentation} was introduced by~\cite{tapnshoot}, and refers to the one-click interactive setting.  One assumes that during image capture, the user taps the touchscreen (once) on the foreground object of interest, from which one can directly segment the object of interest. ~\cite{tapnshoot} uses early fusion; it transforms  the user tap into a guidance map via two shortest-path minimizations and then concatenates the map to the input image.  The authors validate only on simple datasets such as ECSSD~\cite{ecssd} and MSRA10K~\cite{msra}, where the images contain a single dominant foreground object.  As we show later in our benchmarks (see Table~\ref{tab:tapnshoot}), these datasets are so simplistic that properly trained networks with \emph{no} user input can also generate high-quality segmentation masks which are comparable or even surpass the results reported by~\cite{tapnshoot}.  \\

\noindent\textbf{Feature Fusion in Deep Architectures} is an efficient way to leverage complementary information, either from different modalities~\cite{temporal}, or different levels of abstraction~\cite{latematting}. Element-wise multiplication~\cite{guidedprop} and addition~\cite{twostream,nuclei} are two common operations applied for fusing multiple channels. Other strategies include `skip' connections~\cite{fcn}, where features from earlier layers are concatenated with the features extracted from the deeper layers. Recently, a few interactive instance segmentation works have begun exploring outside of the early-fusion paradigm to integrate user guidance~\cite{twostream,guidedprop}. However, these approaches are heavy in their computational footprint, as they increase the number of parameters to be learned by order of hundred of millions~\cite{guidedprop}. Dilution of input information is common-place in deep CNNs as the input gets processed several blocks of convolution~\cite{park2019}. Feature fusion helps preserve input information by reducing the layers of abstraction between the user interaction and the segmentation output.

\section{Proposed Method}
\subsection{Overview}
We follow the conventional paradigm of~\cite{ifcn,itis,majumder19} in which `positive' and `negative' user clicks are transformed into \emph{`guidance'} maps of the same size as the input image.  Unlike~\cite{ifcn,itis,majumder19}, we work within the one-click setting.   The user provides a single `positive' click on the object of interest; this click is then encoded into a single channel guidance map $\mathcal{G}$ (see Sec.~\ref{sub:userinput}). We then feed the $3$-channel RGB image input and the guidance map as an additional channel into a fully convolutional network. Fig.~\ref{fig:fusion}(a) shows an overview of our pipeline. Typically these FCNs are fine-tuned versions of semantic segmentation networks such as FCN-8s~\cite{fcn} or DeepLab~\cite{deeplabv2}.

For our base segmentation network, we use DeepLab-v2~\cite{deeplabv2}; it consists of a ResNet-101~\cite{resnet} feature extraction backbone and a Pyramid Scene Parsing (PSP) module~\cite{psp} acting as the prediction head. Upon receiving the input of size $h \times w \times 4$, the ResNet-101 backbone generates feature maps of dimension $h/8 \times w/8 \times 2048$ (Fig.~\ref{fig:fusion}(a)). %

\subsection{Multi-stage fusion}

\begin{figure}[t]
	\begin{center}
	\includegraphics[width=0.98\linewidth]{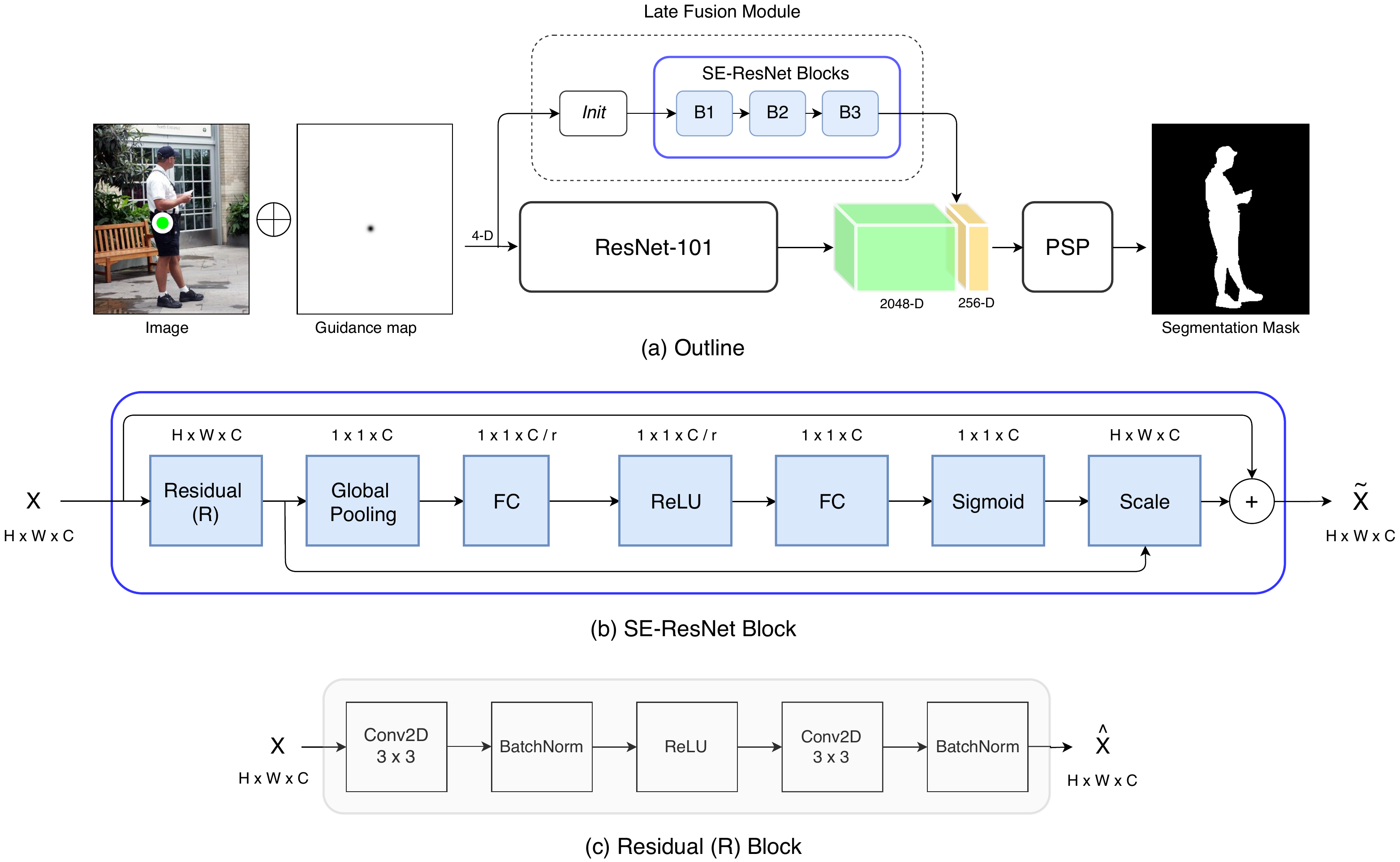}
	\end{center}
	\caption{(a) Overview of our pipeline. Given an  image and a `positive' user click (shown in green circle), we transform the click into a Gaussian guidance map, which is concatenated with the $3$-channel image input and is fed to our segmentation network. For ease of visualization, inverted values for the Gaussian guidance map is shown in the image. The output is the segmentation mask of the selected object. (b) SE-ResNet block (c) Residual block.}
	\label{fig:fusion}
	\vspace{-0.5em}
\end{figure}

The fusion module consists of $3$ \emph{Squeeze-and-Excitation} residual blocks (SE-ResNet)~\cite{senet}. Proposed in~\cite{senet}, SE-ResNet blocks have been shown to effective for a variety of vision tasks such as image classification on ImageNet~\cite{imagenet} and object detection on MS COCO~\cite{mscoco}. SE-ResNet blocks incur minimal additional computational overhead as they consist of two $3\times3$ convolutional layers, two inexpensive fully connected layers and channel-wise scaling operation.

Each SE-ResNet block consists of a \emph{residual} block, a \emph{squeeze} operation which produces a channel descriptor by aggregating feature maps across their spatial operation, dimensionality reduction layer (by reduction ratio \textit{r}) and an \emph{excitation} operation which captures the channel interdependencies. The individual components of the SE-ResNet block is shown in Fig.~\ref{fig:fusion}(b). The residual block consists of two $3\times3$ convolutions, batch normalization, and a ReLU non-linearity (Fig.~\ref{fig:fusion}(c)). We fix the number of filter banks to be $256$ for each of the $3\times3$ convolution. The reduction ratio \textit{r} is kept as 16~\cite{senet}. The input to the fusion block is a $h/4\!\times\!w/4\!\times\!256$ feature map which is obtained by processing the $h\!\times\!w\!\times\!4$ input with $7 \times 7$ convolution operation with stride 2, batch normalization, ReLU non-linearity and a $2 \times 2$ max-pooling operation with stride 2 (\emph{Init} block, Fig.~\ref{fig:fusion}(a)). The final SE-ResNet block downsamples to generate a  $h/8\!\times\!w/8\!\times\!256$ feature map. This is concatenated with the $h/8\!\times\!w/8\!\times\!2048$ obtained from the feature extraction backbone to obtain a $h/8\!\times\!w/8\!\times\!2304$ feature map. 

On top of these feature maps, PSP performs pooling operations at different grid scales on the feature maps to gather the global contextual prior, leading to feature maps of dimensions $h/8\!\times\!w/8\!\times\!512$. The multi-scale feature pooling of PSP~\cite{psp} enables the network to capture objects occurring at different image scales. Pixel-wise foreground-background classification is performed on these down-sampled feature maps. The network outputs a probability map representing whether a pixel belongs to the object of interest or not. Bi-linear interpolation is performed to up-sample the predicted probability map to have the same dimensions as the original input image.

\subsection{Transforming user click}\label{sub:userinput}
In interactive approaches, pixel values of the guidance map are defined as a function of its distance on the image grid to the point of user interaction (Eqn.~\ref{eq:distg}). This includes Euclidean~\cite{ifcn,twostream} and Gaussian guidance maps~\cite{itis}. For each pixel position $\bm{p}$ on the image grid, the pair of distance-based guidance maps for positive ($+$) and negative clicks ($-$) can be computed as 
\begin{equation}\label{eq:distg}
\mathcal{G}^d_{+}(\bm{p}) = \!\min_{\bm{c} \in \{\bm{p}_{+}\}}\!\! d(\bm{p}, \bm{c}) \;\; \text{and} \;\;
\mathcal{G}^d_{-}(\bm{p}) = \!\min_{\bm{c} \in \{\bm{p}_{-}\}}\!\!d(\bm{p}, \bm{c}).
\end{equation}

\noindent For Euclidean guidance maps~\cite{ifcn}, the function $d(\cdot,\cdot)$ is the Euclidean distance. For Gaussian guidance maps, the `min' is replaced by a `max' operator. A more recent approach advocated taking image structures such as super-pixels and region-based object proposals into consideration to generate guidance maps~\cite{majumder19}. To generate the guidance maps, we use Gaussian transformations~\cite{itis} as it offers a favourable trade-off between simplicity and performance. We initialize an image-sized all zero channel and place a Gaussian with a standard deviation of $10$ pixels at the user click location.  Note that we do not use `negative' clicks in our framework.

\subsection{Implementation Details}
\noindent\textbf{Network Optimization.} We train the network to minimize the class-balanced binary cross-entropy loss,
\begin{equation}
    \mathcal{L} = \sum_{j \in N} w_{y_j}\cdot\text{BCE}(y_j, \hat{y}_j)
\end{equation}
where $N$ is the number of pixels in the image, BCE($\cdot$) is the standard cross-entropy loss between the label $y_j$ and the prediction $\hat{y}_j$ at pixel location $j$ given by,
\begin{equation}
    \text{BCE}(y_j, \hat{y}_j) = -y_j\cdot\text{log}\hat{y}_j - (1-y_j)\cdot\text{log}(1-\hat{y}_j)
\end{equation}
$w_{y_j}$ is the inverse normalized frequency of labels $y_j\in\{0, 1\}$ within the mini-batch. We optimize using mini-batch SGD with Nesterov momentum (with default value of $0.9$) and a batch size of 5. The learning rate is fixed at $10^{-8}$ across all epochs and weight decay is $0.0005$. For the ResNet-101 backbone, we initialize the network weights from a model pre-trained on ImageNet~\cite{imagenet}. During training, we first update the early-fusion skeleton for $30$-$35$ epochs. Next we freeze the weights of the early-fusion model and train the late-fusion weights for $5$-$10$ epochs. Finally, we train the joint network for another $5$ epochs.\\

\noindent\textbf{Simulating user clicks.} Manually collecting user interactions is an expensive and arduous process~\cite{benenson19}. In a similar vein as~\cite{tapnshoot} and other interactive segmentation frameworks~\cite{majumder19,ifcn,itis}, we simulate user interactions to train and evaluate our method. During training, we use the ground truth masks of the object instances from the MSRA10K dataset. To initialize, we take the center of mass of the ground truth mask as our user click location; we then jitter the click location by $\mathcal{U}(-50,50)$ pixels randomly. The clicked pixel location is constrained to the confines of the object ground truth mask. %
The random perturbation introduces variation in the training data and also allows better approximation of true user interactions. 

\section{Experimental Validation}
\subsection{Datasets}
We evaluate on six publicly available datasets commonly used to benchmark interactive image segmentation~\cite{tapnshoot,ifcn,itis,majumder19}: MSRA10K~\cite{msra}, ECSSD~\cite{ecssd}, GrabCut~\cite{grabcut}, Berkeley~\cite{berkeley}, PASCAL VOC 2012~\cite{pascal} and MS COCO \cite{mscoco}. We use mean intersection over union (mIoU) of foreground w.r.t. to the ground truth object mask across all instances to evaluate the segmentation accuracy as per existing works~\cite{fcn,ifcn,tapnshoot,itis,majumder19}.

\textbf{MSRA10K} has $10,000$ natural images; the images are characterized by variety in the foreground objects whilst the background is relatively homogeneous. Extended complex scene saliency dataset (\textbf{ECSSD}) is a dataset of $1000$ natural images with structurally complex backgrounds. \textbf{GrabCut} is a dataset consisting of $49$ images with typically a distinct foreground object. It is a popular dataset for benchmarking interactive instance segmentation algorithms. \textbf{Berkeley} dataset consists of $96$ natural images. \textbf{PASCAL VOC 2012} consists of $1464$ training and $1449$ validation images across $20$ different object classes; many images contain multiple objects.   \textbf{MS COCO} is a challenging large-scale image segmentation dataset with $80$ different object categories, $20$ of which are common with the PASCAL VOC $2012$ categories. 

\subsection{Tap-and-Shoot Segmentation}

\begin{table*}[t]
    \setlength{\tabcolsep}{5pt}
	\centering
	\begin{tabular}{lcccccc}
	\toprule
	Method           & $\mathcal{G}$ & res & GrabCut\cite{grabcut} & Berkeley\cite{berkeley} & ECSSD\cite{ecssd}   & MSRA-10K\cite{msra}    \\
	\midrule
	TNS\cite{tapnshoot} & \small{\Checkmark}  & 256 & 72.3 / 79.0  & 55.7 / 67.0 & 70.3 / 76.0 & 81.1 / 85.0    \\
	\hdashline
	\textit{vgg-baseline} & \small{\XSolidBrush}  & 256  & 73.5 / 77.4 &  58.2 / 63.2 & 71.2 / 72.3 &  83.4 / 86.2 \\
	\textit{vgg-early} & \small{\Checkmark}  & 256  & 76.2 / 80.1  & 62.8 / 65.3   & 74.8 / 76.5 &  87.1 / 87.5  \\
	\hdashline
	\textit{resnet-baseline}  & \small{\XSolidBrush} & 256    &  81.6 / 83.0  & 68.5 / 68.2    &  80.2 / 82.0  & 86.4 / 86.9 \\
		\textit{resnet-early}     & \small{\Checkmark}   & 256    & 83.3 / 84.3  & 75.0 / 75.3 & 82.0 / 83.6 & 88.6 / 89.6  \\
			\textit{resnet-multi}    & \small{\Checkmark}   & 256    & 84.1 / 85.7  & 75.1 / 78.4  & 81.9 / 85.2 & 91.5 / 92.1  \\
		\hdashline
	\textit{resnet-baseline}  & \small{\XSolidBrush} & 512    & 76.1 / 79.0  & 65.5 / 68.3 & 79.9 / 82.6   & 87.0 / 87.9 \\
	\textit{resnet-early}     & \small{\Checkmark}   & 512    & 82.9 / 84.5  & 76.2 / 78.1  & 85.6 / 85.7 & 91.5 / 91.4  \\
	\textit{resnet-multi}    & \small{\Checkmark}   & 512    & 83.1 / 86.2  & 80.1 / 81.3  & 86.8 / 87.1 & 92.5 / 93.1  \\
 	\midrule
	\bottomrule
	\end{tabular}
	\vspace{0.5em}
	\caption{\textbf{Ablation Study: Tap-and-Shoot Segmentation}. `res' refers to the image resolution used during training. We report average mIoU for the segmentation results after training for 16K iterations and after training convergence. The ~\textit{-baseline} models receive a $3$-channel RGB image as input without the guidance map $\mathcal{G}$.}%
	\label{tab:tapnshoot}
\end{table*}
Following~\cite{tapnshoot}, we use MSRA10K~\cite{msra} for training and partition the dataset into three non-overlapping subsets of $8000$, $1000$ and $1000$ images as our training, validation and test set. We report the mIoU after training for 16K iterations and again after network convergence (at 43k iterations for us, vs. 260k iterations in~\cite{tapnshoot}) in Table~\ref{tab:tapnshoot}. During training, we resize the images to $512\times 512$ pixels. This choice of resolution is driven primarily by matching the resolution to that of the training images for the ResNet-101 backbone~\cite{resnet}. 

The \textit{-baseline} models are trained using only the $3$-channel RGB image and the instance ground truth mask without any user click transformations. The~\textit{-early} models use Gaussian guidance maps~\cite{itis}; the network input is $3$-channel RGB image and Gaussian encoding of the user's tap on the object of interest (Fig.~\ref{fig:motivation}(a)). The \textit{-multi} models refer to the multi-stage fusion models with Gaussian encoding of user clicks. Note that we do not train a late-fusion model; standalone late-fusion models show inferior performance compared to their early-fusion counterparts~\cite{guidedprop}.

From Table.~\ref{tab:tapnshoot}, we observe that our trained network converges mostly within 16K iterations. For simplistic datasets such as MSRA10K and ECSSD, the \textit{vgg-baseline} without user click transformation compares favourably with the approach of~\cite{tapnshoot} at the same training resolution of $256\times 256$. \textit{resnet-baseline} models trained with $512 \times 512$ images significantly outperform~\cite{tapnshoot} reporting absolute mIoU gains of till $7\%$ across the datasets.  Based on this result alone, we conclude that one-click (and standard) interactive segmentation approaches should be benchmarked on more challenging datasets.  Examples include PASCAL VOC 2012 and MS COCO, which feature cluttered scenes, multiple objects, occlusions and challenging lighting conditions. (see Table~\ref{tab:soa}).

Furthermore, with only the Gaussian transformation and ResNet-101 backbone trained on $512\times 512$, we are able to achieve mIoU increase in the range of $5$-$11\%$ across datasets at convergence \emph{w.r.t}~\cite{tapnshoot}. Having the multi-stage fusion offers us absolute mIoU gains of $1$-$4\%$ w.r.t the early fusion variant (\emph{resnet-early vs. resnet-multi} when trained with $512 \times 512$ images). Additionally, our \textit{resnet} models require significantly less memory; $195.8$ MB (stored as $32$-bit/$4$-byte floating point numbers) instead of the $652.45$ MB required for the segmentation network of~\cite{tapnshoot}.

\subsection{Interactive image segmentation}
Approaches in the literature~\cite{ifcn,itis,majumder19,twostream} are typically evaluated by (1) the average number of clicks needed to reach the desired level of segmentation ($@85\%$ mIoU for PASCAL VOC $2012$, MS COCO, $@90\%$ mIoU for the less challenging Grabcut and Berkeley) and (2) the average mIoU \textit{vs} the number of clicks.

\begin{figure}[t]
	\begin{center}
	\includegraphics[width=1\linewidth]{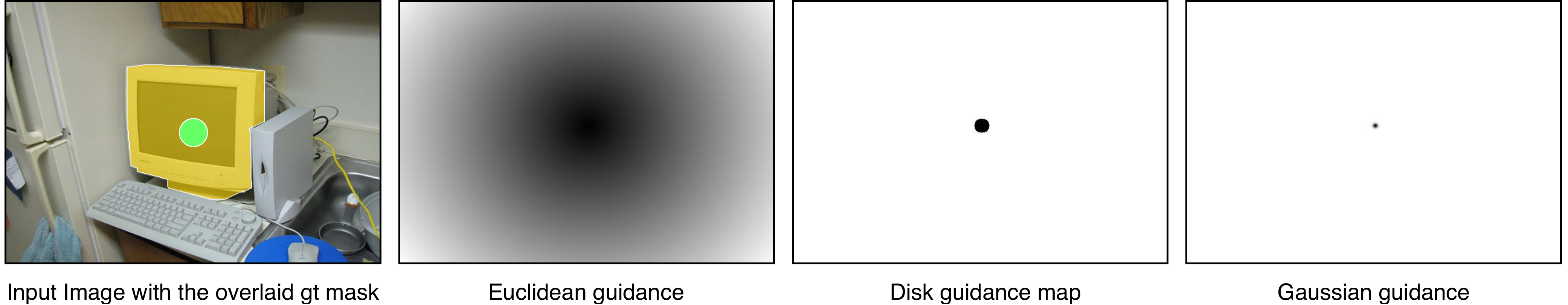}
	\end{center}
	\caption{\textbf{Examples of guidance maps.} Given a click (shown as green circle) on the object  of interest, existing approaches transform it into guidance maps and uses it as an additional input channel. For ease of visualization, inverted values for the disk guidance map and the Gaussian guidance map are shown in the image.}
	\label{fig:exampleguidance}
\end{figure}

The first criterion is primarily geared towards annotation tasks~\cite{itis,majumder19} where high-quality segments are desired for each instance in the scene; the fewer the number of clicks, the lower the annotation effort. In this work, we are concerned primarily with achieving high-quality segments for the object of interest given only a single click. Accordingly, given a single user click, we report the average mIoU across all instances for the GrabCut, Berkeley and the PASCAL VOC $2012$ \textit{val} dataset. For MS COCO object instances, following~\cite{ifcn}, we split the dataset into the $20$ PASCAL VOC $2012$ categories and the $60$ additional categories, and randomly sample $10$ images per category for evaluation. We also report the average mIoU across the sampled $800$ MS COCO instances~\cite{twostream}. 

For training~\cite{ifcn,majumder19,twostream}, we use the ground truth masks of object instances from PASCAL VOC $2012$~\cite{pascal} \textit{train} set with additional masks from Semantic Boundaries Dataset (SBD)~\cite{sbd} resulting in $10582$ images. Note that unlike~\cite{itis}, we do not use the training instances from MS COCO. \\

\noindent\textbf{Ablation Study.} We perform extensive ablation studies to thoroughly analyze the effectiveness of the individual components of our one-click segmentation framework. First, to validate our choice of guidance maps, we consider the user click transformations commonly used in existing interactive segmentation algorithms - Euclidean distance maps~\cite{ifcn,twostream}, Gaussian distance maps~\cite{itis} and disk~\cite{benenson19}. Fig.~\ref{fig:exampleguidance} shows examples of such guidance maps. For each kind of guidance map, we train separate networks to understand the impact of different user click transformations. For evaluation, we report the average mIoU over all instances in the dataset, given a single click (see Table~\ref{tab:ablation}). Next, we study the impact of our proposed late-fusion module (denoted by \textit{-multi} in Table.~\ref{tab:ablation}); we observe an average mIoU improvement of around $1.8\%$ across different datasets. \\

\begin{table*}[t]
    \setlength{\tabcolsep}{5pt}
	\centering
	\begin{tabular}{llllll}
	\toprule
	$\mathcal{G}$ & GrabCut & Berkeley & VOC12   & COCO-20     & COCO-60 \\
	\midrule
    Euclidean~\cite{ifcn}       & 82.6    & 82.7     & 75.1    & 63.2        & 46.8  \\
    Disk~\cite{benenson19}     & 84.5    & 81.3     & 74.5    & 65.3        & 51.5 \\
	\textbf{Gaussian}~\cite{itis}         & 84.0      & 82.9     & 78.1    & 64.2        & 49.8 \\  \hdashline
	\textbf{Gaussian}-\emph{multi} & \textbf{86.2}{\tiny ($2.2\uparrow$)} & \textbf{84.0}{\tiny ($1.1\uparrow$)} & \textbf{80.8}{\tiny ($2.7\uparrow$)} & 64.5{\tiny ($0.3\uparrow$)} & \textbf{52.3}{\tiny ($2.5\uparrow$)} \\
	\midrule
	\bottomrule
	\end{tabular}
	\vspace{0.5em}
	\caption{\textbf{User Click Transformation.} The best results are indicated in \textbf{bold}. COCO-$20$ and COCO-$60$ refers to the instances from $20$ overlapping categories and $60$ non-overlapping categories of PASCAL VOC $2012$ respectively.}
	\label{tab:ablation}
\end{table*}

\begin{figure}[t]
    \begin{center}
	\includegraphics[width=0.95\linewidth]{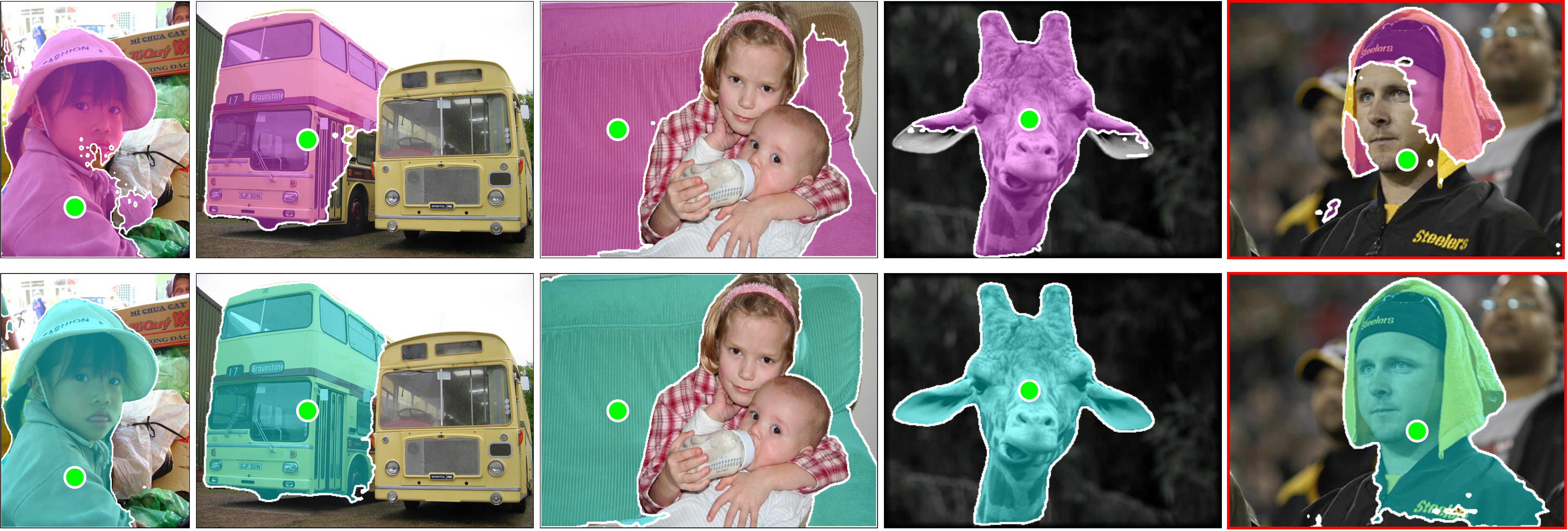}
    \end{center}
    \caption{\textbf{Qualitative Results.} Incorporating the user clicks at different stages of the network leads to an improvement in the quality of masks generated (second row) \emph{w.r.t} the early-fusion variants (first row). Click locations are shown in green circles. The extreme right column shows a scenario where both the networks failed to generate a satisfactory mask.}
	\label{fig:qualitative}
\end{figure}

\noindent\textbf{One-click segmentation.} We compare the segmentation performance of our method with existing interactive instance segmentation approaches (see Table~\ref{tab:soa}). The approaches are grouped separately into $3$ different categories - pre-deep learning approaches, deep learning-based interactive instance segmentation approaches and tap-and-shoot segmentation approaches. From Table.~\ref{tab:soa}, we observe that our approach outperforms the classical interactive segmentation works by a significant margin reporting $40\%$ absolute improvement in average mIoU. We also outperform existing state-of-the-art interactive instance segmentation approaches~\cite{majumder19,itis} by a considerable margin ($>3\%$). Additionally, we report an absolute mIoU improvement of $7.2\%$ and $17\%$ on Grabcut and Berkeley over the tap-and-shoot segmentation framework of~\cite{tapnshoot}.  We show qualitative results to demonstrate the effectiveness of our proposed algorithm (see Fig.~\ref{fig:qualitative}). The resulting segmentations demonstrate that our approach is highly effective for the one-click segmentation paradigm. 

\section{User Study}
Across existing state-of-the-art interactive frameworks~\cite{ifcn,itis,majumder19}, user clicks are simulated following the protocols established in~\cite{ifcn,itis}. For our user study, we consult $5$ participants uninitiated to the task of interactive segmentation. We prepare a toy dataset with $50$ object instances from the MSRA10K~\cite{msra} dataset. We presented the image with the segmentation mask for the target object overlaid on the image and asked the users to provide their click.

\begin{table*}[t]
    \setlength{\tabcolsep}{3.2pt} 
	\centering
	\begin{tabular}{lllllll}
	\toprule
	Method       & Network & GrabCut & Berkeley & VOC12   & COCO-20 & COCO-60\\ 
	\midrule
	GC\cite{graphcuts} & - & 41.7  & 33.8  & 27.7     & -          & 8.9         \\
	GM\cite{geodesicmatting} & -  & 23.7  & 24.5 & 23.8 & -             & 22.1         \\
    GD\cite{geodesic} & - & 48.8  & 36.1 &  31.0      & -             & 25.2         \\
	\hdashline
    iFCN\cite{ifcn} & FCN-8s\cite{fcn}  & 62.9     & 61.3         & 53.6       &             &  \textbf{42.9}       \\
	ITIS\cite{itis} & DeepLabv3+\cite{deeplabv3} & 82.1        &  -       & 71.0      & -         & -     \\ 
	CAG\cite{majumder19} & FCN-8s\cite{fcn} & \textbf{83.2}      &   -       &  \textbf{74.0}       & -            & - \\ 
	TS\cite{twostream} & FCN-8s\cite{fcn}  & 77.7  & \textbf{74.5} & 62.3      & \textbf{42.5}        & 42.5          \\
	\hdashline
	TNS\cite{tapnshoot} & FCN-8s\cite{fcn} & 79.0  & 67.0 & - & - & - \\
	\hline
	\textit{Ours-best} & DeepLabv2\cite{deeplabv2} & \textbf{86.2}{\tiny ($3.0\uparrow$)} & \textbf{84.0}{\tiny ($9.5\uparrow$)} & \textbf{80.8}{\tiny ($6.8\uparrow$)} & \textbf{64.5}{\tiny ($22.0\uparrow$)} & \textbf{52.3}{\tiny ($9.6\uparrow$)} \\
	\midrule
	\bottomrule
	\end{tabular}
	\vspace{0.5em}
	\caption{\textbf{Average mIoU given a single click.} The approaches are grouped separately into $3$ different categories - pre-deep learning approaches, deep learning-based interactive instance segmentation approaches and tap-and-shoot segmentation approaches respectively. For GC\cite{graphcuts}, GM\cite{geodesicmatting}, GD\cite{geodesic}, and iFCN\cite{ifcn} we make use of the values provided by the authors of iFCN\cite{ifcn}. The mIoU improvement (in \%) over existing state-of-the-art approaches is indicated using~$\uparrow$.}
	\label{tab:soa}
\end{table*}

During training the object selection stage, we applied random perturbations of $\mathcal{U}(-50,50)$ pixels to the center of mass of the object instance to obtain the final user click. Our user study found that participants placed clicks at a mean distance of $24$ pixels from the center of the mask with a standard deviation of $27$ pixels. This result validates our assumption that users are more likely to click in the vicinity of the object's center-of-mass.  It also supports our click sampling scheme for generating training instances when training the object selection stage. On average, we observed that users took $2.3$ seconds with a standard deviation of 0.8 seconds to position their click. 

\section{ Conclusion}
In this work, we propose a one-click segmentation framework that produces high-quality segmentation masks. We validated our design choices through detailed ablation studies;  we observed that having a multi-stage module improves the segmentation framework and gives the network an edge over its early-fusion variants. Via experiments, we observed that for the single click scenario, our proposed approach significantly outperforms existing state-of-the-art approaches - including the more complicated interactive instance segmentation models using state-of-the-art segmentation models~\cite{deeplabv3}. %

However, we observe existing tap-and-shoot segmentation frameworks~\cite{tapnshoot}, including our proposed framework, are limited by their inability to learn from negative clicks~\cite{ifcn,majumder19,itis}. One major drawback of such a training scenario is that the network does not have a notion of corrective clicking; if the generated segmentation mask extends beyond the object boundaries, it cannot rectify this mistake. Clicking on locations outside the object can mitigate this effect, though this then deviates from tap-and-shoot interaction. 
\\
\\
\noindent\textbf{Acknowledgment.} This work was supported in part by National Research Foundation Singapore under its NRF Fellowship Programme [NRF-NRFFAI1-2019-0001] and NUS Startup Grant R-252-000-A40-133.

\bibliographystyle{unsrt}
\bibliography{references}

\end{document}